\documentclass[letterpaper, 10 pt, conference]{ieeeconf}  

\IEEEoverridecommandlockouts                              

\makeatletter

\newcommand{\Rmnum}[1]{\expandafter\@slowromancap\romannumeral #1@}
\makeatother
\makeatletter
\let\NAT@parse\undefined
\makeatother
\usepackage[pagebackref=false,breaklinks=true,colorlinks=true,bookmarks=false,linkcolor={red!50!black},urlcolor={blue},citecolor={green!60!black}]{hyperref} 
\usepackage{graphics} 
\usepackage{epsfig} 
\usepackage{mathptmx} 
\usepackage{times} 
\usepackage{amsmath} 
\usepackage{amssymb}  
\usepackage{graphicx}
\usepackage{wrapfig}
\usepackage{xcolor}

\title{\LARGE \bf
Towards Intuitive Human–Robot Interaction through Embodied Gesture-Driven Control with Woven Tactile Skins
}
\usepackage{booktabs}
\usepackage{subcaption}
\usepackage{balance}

\author{ChunPing Lam$^{1,2*}$, Xiangjia Chen$^{1,3*}$, Chenming Wu$^{3}$, Hao Chen$^{1}$, Binzhi Sun$^{1,2}$, \\ Guoxin Fang$^{1,2}$, Charlie C.L. Wang$^{4}$,  Chengkai Dai$^{1,3\dagger}$, Yeung Yam$^{1,2,3\dagger}$
\thanks{$^{*}$ Joint first authors.}%
\thanks{$^{1}$ Centre for Perceptual and Interactive Intelligence (CPII), Hong Kong SAR, China.}%
\thanks{$^{2}$ Department of Mechanical and Automation Engineering, The Chinese University of Hong Kong, Hong Kong SAR, China.}
\thanks{$^{3}$ Texense Ltd., Hong Kong SAR, China.}%
\thanks{$^{4}$ Department of Mechanical, Aerospace and Civil
Engineering, University of Manchester, United Kingdom.}
\thanks{$^{\dagger}$ Joint corresponding authors ({\tt\small yyam@mae.cuhk.edu.hk}, {\tt\small ckdai@cpii.hk}).} %
}

\begin{document}
\maketitle
\thispagestyle{empty}
\pagestyle{empty}

\begin{abstract}

This paper presents a novel human–robot interaction (HRI) framework that enables intuitive gesture-driven control through a capacitance-based woven tactile skin. Unlike conventional interfaces that rely on panels or handheld devices, the woven tactile skin integrates seamlessly with curved robot surfaces, enabling embodied interaction and narrowing the gap between human intent and robot response. Its woven design combines fabric-like flexibility with structural stability and dense multi-channel sensing through the interlaced conductive threads. Building on this capability, we define a gesture–action mapping of 14 single- and multi-touch gestures that cover representative robot commands, including task-space motion and auxiliary functions. A lightweight convolution–transformer model designed for gesture recognition in real time achieves an accuracy of near-100\%, outperforming prior baseline approaches. Experiments on robot arm tasks, including pick-and-place and pouring, demonstrate that our system reduces task completion time by up to 57\% compared with keyboard panels and teach pendants. Overall, our proposed framework demonstrates a practical pathway toward more natural and efficient embodied HRI.
\end{abstract}

\section{Introduction}

Human–robot interaction (HRI) is central to the deployment of robotic systems in daily life and collaborative settings \cite{goodrich2008human}. For robots to operate effectively alongside humans, the interaction interfaces must be intuitive and user-friendly. Conventional interfaces—such as control panels, handheld devices, or teach pendants—often create a gap between human intent and robot response, thereby hindering natural collaboration \cite{Iskandar2024ScienceRobotics}. 

Tactile interfaces, also known as robot skins, offer a promising alternative by enabling direct physical interaction with the robot. Tactile skins that can be seamlessly integrated onto curved robot surfaces unlock the possibility of on-surface interaction, where users can engage with the robot body in an intuitive and embodied manner \cite{jiang2025robot}. However, most existing tactile skins respond only to direct force inputs, typically translating touch into compliance control or safety behaviors \cite{luo2025tactileroboticsoutlook}. While this suffices for basic collaboration, it limits interaction bandwidth and cannot interact with richer contact patterns. Gestures, in contrast, can be viewed as structured spatiotemporal interactions that evolve dynamically across time, space, and multiple contact points \cite{taunyazov2020tactile}. These patterns extend beyond individual points of contact, offering a richer vocabulary for users to express their intentions. By interpreting tactile input as gestures, robotic skins can evolve from basic responsiveness to more intuitive, high-level interaction interfaces.

\begin{figure}[t]
    \hfill
    \includegraphics[width=0.48\textwidth]{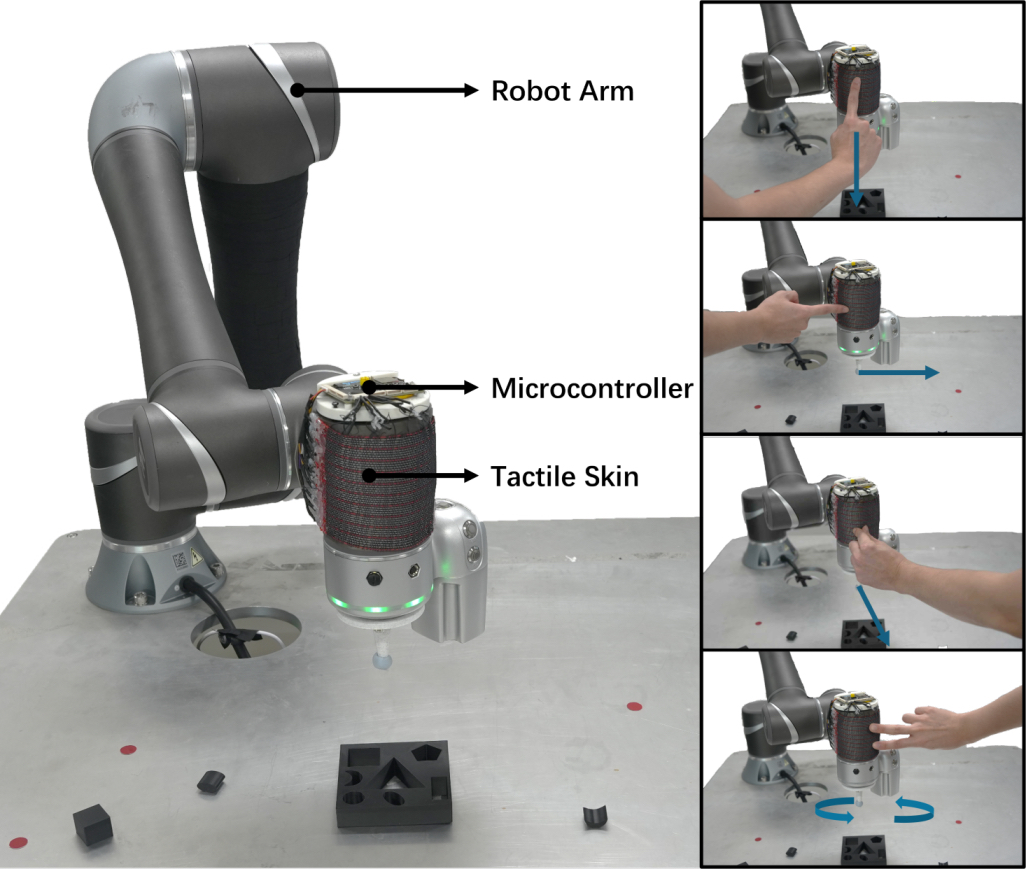} 
    \caption{Framework overview: A woven tactile skin is integrated onto the robot’s surface to enable embodied, gesture-driven control (left). Gestures made on the skin are detected and translated into robot actions, facilitating intuitive human-robot interaction (right).}
    \label{fig:system_overview}
\end{figure}

Despite this potential, gesture-based tactile interaction on robot surfaces remains challenging. One difficulty lies in sensing: most existing tactile skins are fabricated either as flexible films \cite{Taunyazov21}, which are difficult to scale up for high-resolution sensing and large-area coverage, or as stretchable fabrics \cite{robosweater}, which often suffer from mechanical instability when deployed on curved robot surfaces. Another challenge lies in gesture design: to ensure truly intuitive interaction, gesture sets must naturally align with user intent and effectively represent common robot commands without causing ambiguity \cite{niemi2021user}. Moreover, achieving robust recognition remains challenging -- while vision-based methods have made significant progress in understanding gestures, tactile sensing still lacks mature techniques for directly capturing continuous spatiotemporal patterns on robot surfaces \cite{VERMA2025100284}.

To tackle these challenges, we propose a novel human–robot interaction framework for intuitive on-surface control, centered around a capacitance-based woven tactile skin. Its interlaced structure offers structural stability and inherently supports robust, multi-channel sensing of dynamic, multi-touch interactions. Meanwhile, its fabric-like flexibility enables seamless conformity to robot surfaces. Building on this foundation, we introduce a gesture vocabulary comprising 14 dynamic single- and multi-touch patterns—such as swipes, pushes, pinches, and circular strokes involving 1 to 5 contact points. These gestures are directly mapped to robot actions, including translations and rotations in task space, as well as auxiliary functions, forming an intuitive and expressive command set for interaction. To facilitate gesture recognition from tactile input, we design a hybrid architecture that integrates a lightweight convolutional stem for per-frame spatial encoding with a temporal transformer to capture long-range dependencies across frames, enabling accurate and low-latency recognition in real-world scenarios.

We further validate the proposed framework in real-world, vision-free settings through pick-and-place and pouring tasks, where robot actions are directly controlled by human gestures performed on the robot’s surface. Compared to conventional HRI interfaces like keyboard panels or teach pendants, our approach reduces task completion time by 23–57\%, demonstrating the intuitiveness and efficiency of gesture-driven control.

The main contributions of this work are summarized as follows:
\begin{itemize}
    \item We introduce a novel embodied gesture-driven human–robot interaction framework using a capacitance-based woven tactile skin, where the interlaced woven structure ensures stable sensing of dynamic and multi-touch tactile inputs such as human gestures (Section~\ref{sec:woven_sensor}).
    \item We design a gesture vocabulary of 14 single- and multi-touch patterns with up to five contact points, which are intuitively mapped to representative robot commands, including task-space translation, rotation, and essential auxiliary control functions that improve operational efficiency (Section~\ref{sec:action_mapping}).
    \item We propose a hybrid convolution–transformer recognition model for classifying gestures from spatiotemporal tactile signals, achieving a near-100\% success rate in our tasks and outperforming LSTM-based baselines, while maintaining low latency for real-time human–robot interaction (Section~\ref{sec:gesture_recognition}).

\end{itemize}
\section{Related Work}
\label{sec:related_work}


\subsection{Robot Skin}
The rapidly advancing robotic technique extended its focus beyond motion control, in particular, the research on tactile sensing enables robots to ``feel'' their environment and adjust their movements~\cite{1984, lee1999tactile}. Early works mechanically mounted discrete pressure sensors onto robot exteriors~\cite{nicholls1989survey}. However, manual calibration, complex assembly, low sensor density, and cumbersome wiring limited practical adoption. The concept of electronic skin (e-skin) advanced the field by enabling conformal, flexible sensor sheets for precise tactile localization on curved surfaces~\cite{flexible2004, stretchable_skin_ref_schwartz2013}. Yet, such designs often suffered from mechanical fatigue and high costs due to custom fabrication for each geometry.

Textile-based tactile sensors have recently emerged as a promising alternative, utilizing 3D weaving or knitting techniques with conductive yarns to form scalable, flexible tactile grids~\cite{stassi2014flexible}. Each intersecting point acts as a tactile sensing node, and the woven structure offers low-cost fabrication and ease of integration, akin to robotic ``garments''~\cite{robosweater}. Engineering reliable and uniform sensing grids remains a challenge, but advances in thread path design and weaving technology have improved mechanical stability and signal quality~\cite{tamura2020woven}.

Two main tactile sensing modalities are seen in the literature: resistive and capacitive. Resistive sensors correlate applied force with electrical resistance but often face high power consumption and signal drift~\cite{dahiya2010tactile}. Capacitive sensors, known for high sensitivity and low power usage, are widely used in smartphone touchscreens and now in robotic applications~\cite{hu2014flexible}. However, their integration onto robot surfaces is complicated by susceptibility to electromagnetic interference, necessitating stable electrode architectures. Recent work exploits woven conductive threads as stable electrodes, leveraging the mechanical interlocking of warp and weft for robustness and reliable multimodal sensing~\cite{tamura2020woven, stassi2014flexible}.

\subsection{Gesture-based HRI Interfaces}
Traditional HRI interfaces, such as control panels, often demand considerable learning efforts for effective human-robot interaction~\cite{goodrich2008human}. In contrast, gesture-driven control facilitates intuitive, embodied interaction, effectively bridging the gap between user intent and robot response~\cite{billard2019trends}. 

Research indicates that robot-initiated gestures, such as intra-hugs, foster understanding and comfort~\cite{block2023arms}, while playful interactions like hand-clapping games strengthen the human-robot connection~\cite{fitter2016qualitative}. Despite the exploration of these interaction modalities, tactile gesture recognition for dynamic interactions remains relatively under-researched~\cite{yuan2017gelsight}. Notably, \cite{liu2022machine} emphasizes that multi-touch and multi-finger gestures can broaden the vocabulary of robot commands. Recent findings~\cite{crowder2025social} reveal that tactile gestures on a robot's fabric surface can effectively convey social cues, enhancing engagement and collaboration. These insights highlight the promise of tactile gesture recognition as a natural and effective interface for HRI.

\subsection{Tactile Gesture Recognition}
To interpret the spatiotemporal dynamics of tactile gesture data, machine learning approaches have been widely adopted. Traditional recurrent neural networks (RNNs), such as long short-term memory (LSTM) models, capture temporal dependencies but typically flatten spatial information, resulting in a loss of critical structure~\cite{graves2012long, pastor2020bayesian, bai2022tactile}. Convolutional neural networks (CNNs) are well-suited for extracting spatial features from tactile frames, while transformer-based architectures enable efficient modeling of long-range temporal dependencies~\cite{awan2023predicting, vaswani2017attention}. Furthermore, self-supervised pre-training strategies on large-scale, unlabeled tactile data have shown to enhance feature extraction and generalization, leading to improved classification performance and faster convergence during fine-tuning~\cite{he2020momentum, dave2024multimodal}. While significant progress has been made in tactile gesture recognition and spatiotemporal modeling, there remains a gap in methods and datasets specifically tailored for robot-embodied, gesture-driven control with conformal tactile skins---the focus of our present work.

\section{Woven-Based Tactile Skin}
Our novel woven tactile skin, fabricated via a weaving process, detects tactile interactions through variations in capacitance between interwoven conductive threads. The interlaced architecture provides mechanical stability and intrinsic multi-channel scalability, enabling reliable large-area integration on curved robot surfaces. Below, we provide a detailed description of the skin.

\label{sec:woven_sensor}
\subsection{Skin Design with Woven Structure}
\begin{figure}[t]
    \centering
    \includegraphics[width=0.45\textwidth]{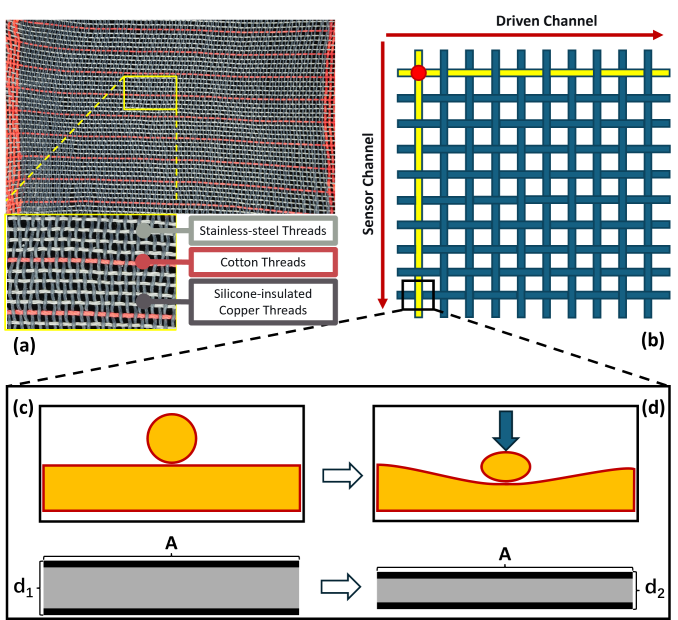} 
    \caption{(a) The tactile skin consists of a capacitive sensing grid formed by interlacing capacitive threads; red cotton threads are used only for channel size indication. (b-d) Applied force reduces electrode spacing and alters overlap area, producing capacitance changes. }
    \label{fig:sensor_design}
\end{figure}

The tactile skin is fabricated on a woven textile structure, where interlaced warp and weft threads provide the underlying framework for a distributed capacitive sensing system. Within this woven matrix, two distinct categories of conductive threads are strategically integrated: silicone-insulated copper wires, which function as positive electrodes, and stainless-steel threads, which serve as negative electrodes, as depicted in Fig.~\ref{fig:sensor_design}(a). During the weaving process, the positioning of these threads is carefully controlled to maintain uniform spacing, alignment, and mechanical stability, ultimately forming a capacitive grid across the fabric. In this configuration, the conductive threads act as opposing plates of multiple miniature capacitors, while the woven textile medium and the surrounding air jointly operate as the dielectric layer. Such an arrangement effectively transforms the fabric into a dense, high-resolution sensing array capable of capturing subtle variations in applied pressure and contact area. Beyond its sensing functionality, this structural design imparts a number of desirable physical properties: the textile remains lightweight and highly flexible, enabling it to be seamlessly integrated onto non-planar and curved robot surfaces, as illustrated in Fig.~\ref{fig:sensor_design}(b).

\subsection{Working Principle for Tactile Sensing}
For tactile sensing, applied force compresses the woven layers and deforms the silicone-insulated fibers. This structural deformation alters the electrode separation distance and the effective overlap area, leading to measurable capacitance variations. As a result, the skin can capture both static and dynamic tactile events, ranging from light touches to continuous pressing. Fig.~\ref{fig:sensor_design}(c) illustrates the undeformed baseline state, while Fig.~\ref{fig:sensor_design}(d) shows deformation under contact and the corresponding capacitance change. Based on this working principle, we can detect stresses up to 100 kPa.

\subsection{Skin Fabrication and Signal Readout}
\subsubsection*{Fabrication}
We fabricate the tactile skin using a customized 3D weaving system that enables independent control of warp and weft threads using stainless-steel threads and silicone-insulated copper threads, respectively \cite{CHEN2024102819}. In our design, 51 warp threads, with 40 conductive and 11 cotton, and 80 weft threads are woven with a spacing of approximately 2.0 mm $\times$ 1.75 mm, resulting in a sensing area of about 100.0 mm $\times$ 140.0 mm.
\vspace{5px}

\subsubsection*{Signal Readout}
Every group of 4 adjacent warp threads and 8 adjacent weft threads is grouped to define a single capacitive sensing channel, resulting in a 10 $\times$ 10 grid (100 sensing channels) in total. Capacitance values from this sensing grid are acquired at 200 Hz using an AD5941 chip with a parallel four-phase multiplexing strategy, in which positive and negative channels are divided into A/B groups and each scan completes within 1 ms. The acquired data are then transmitted by an ESP32-S3 microcontroller (shown in Fig.~\ref{fig:system_overview}) to a PC via serial communication for further processing and analysis. To ensure reliable tactile input, a moving average filter with 100 ms latency is applied, and all signals are referenced to a baseline established during a 500 ms static calibration period.

\section{Gesture–Action Mapping }
\label{sec:action_mapping}
\begin{figure}
\centering
\includegraphics[width=0.45\textwidth]{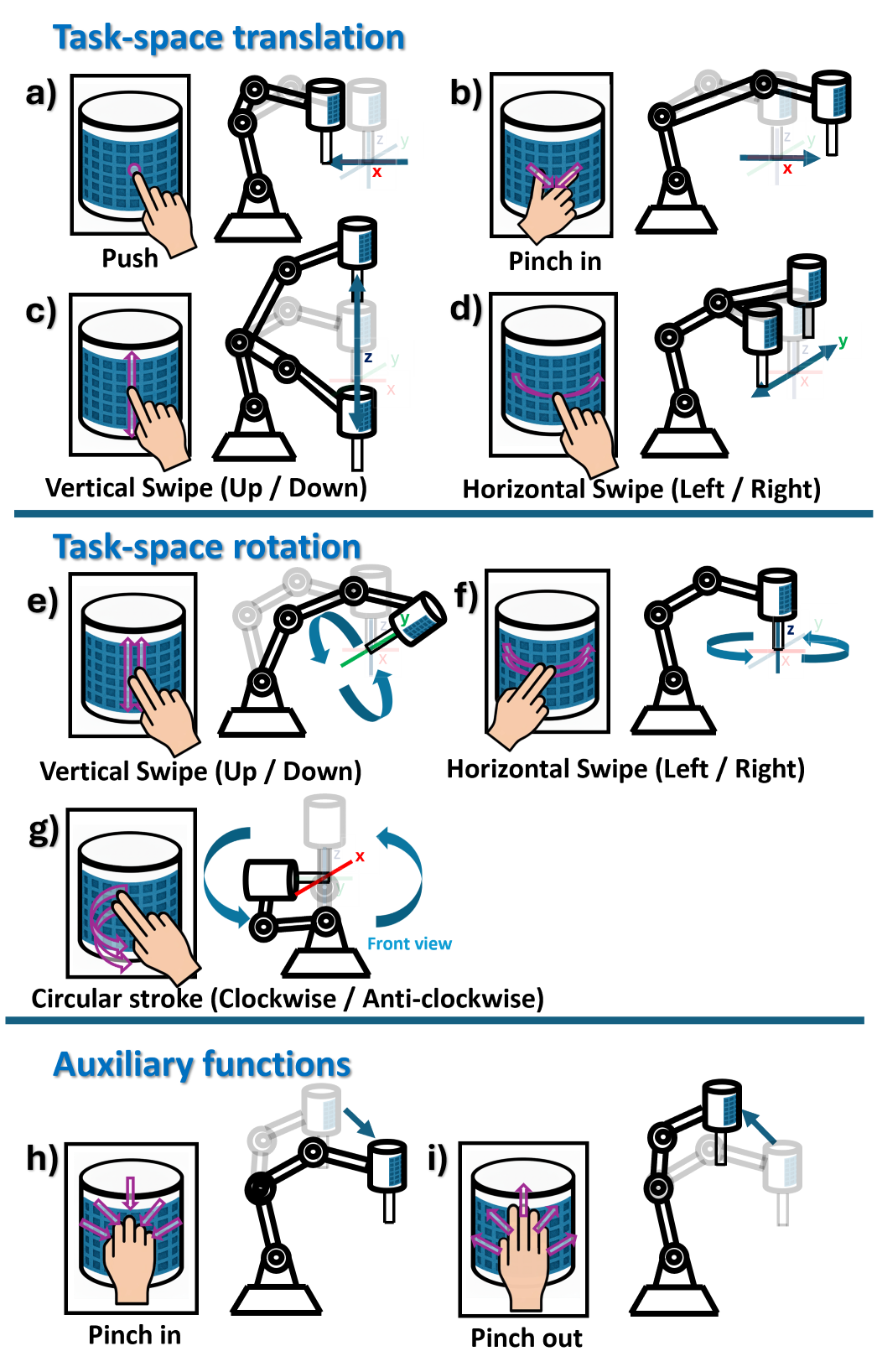} 
\caption{We design an intuitive gesture–action mapping that enables gesture-driven control through tactile skin attached to the robot arm's end link. 14 gestures are organized into three categories: (a–d) task-space translation ($x$, $y$, $z$ axes via push, pinch-in, and single-finger swipes), (e–g) task-space rotation ($x$, $y$, $z$ axes via two-finger swipes and circular strokes), and (h–i) auxiliary functions (five-finger pinch-in/out for initial pose and home posture).
}
\label{fig:gesture_design}
\end{figure}

To enable intuitive human–robot interaction, we develop a gesture vocabulary of 14 dynamic single- and multi-touch patterns, illustrated in Fig.~\ref{fig:gesture_design}, derived from empirical evaluation of practical human–robot collaboration tasks (e.g., pick-and-place). The tactile skin is mounted on the robot arm’s end link, positioning the interface directly at the site of manipulation and ensuring that gestures correspond naturally to task execution. Based on this placement, the gestures are systematically mapped to robot actions, covering task-space translation, rotation, and auxiliary commands. The design follows two principles:

\begin{enumerate}
\item \textbf{Naturalness:} gestures should align with human intuition, minimizing the cognitive load required for learning;
\item \textbf{Completeness:} the vocabulary should cover representative robot commands—translation and rotation in task space (i.e., the end-effector coordinate frame)—together with auxiliary functions that enable safe recovery from undesired states and improve efficiency in repetitive tasks.
\end{enumerate}
On this basis, the gesture vocabulary is organized into three categories: task-space translation (6 gestures), task-space rotation (6 gestures), and auxiliary functions (2 gestures).

\emph{Task-space translation:} A single-finger vertical swipe maps to motion along the $z$-axis, with upward and downward swipes corresponding to positive and negative directions. Similarly, a single-finger horizontal swipe controls motion along the $y$-axis, where rightward and leftward swipes correspond to positive and negative directions. Notice that $x$-axis control is realized through a combination of single- and two-finger gestures: forward translation is expressed by a two-finger pinch in that conveys grasping the surface and moving it forward, whereas backward translation is performed by a single-finger push, intuitively corresponding to pushing the robot away. These six gestures for translation are illustrated in Fig.~\ref{fig:gesture_design}(a–d).

\emph{Task-space rotation:} Two-finger vertical swipes control rotation around the $y$-axis, with upward and downward swipes corresponding to positive and negative rotation, respectively. Similarly, two-finger horizontal swipes control rotation around the $z$-axis, where leftward swipes indicate negative rotation and rightward swipes indicate positive rotation. Rotation around the $x$-axis is expressed by a two-finger circular stroke, where clockwise and anti-clockwise strokes intuitively encode opposite directions. These six gestures for rotation are illustrated in Fig.~\ref{fig:gesture_design}(e–g).

\emph{Auxiliary functions:} A five-finger pinch is reserved for recovery and task initialization, where its high distinctiveness ensures a very low likelihood of accidental activation. An inward pinch, resembling a “release-in” motion, returns the robot to the task’s initial pose, enabling rapid re-execution from a starting configuration, while an outward pinch, resembling a “release-out” motion, moves the robot to its home posture for safe resetting. The two auxiliary gestures are illustrated in Fig.~\ref{fig:gesture_design}(h) and (i).

\vspace{5px}
We also highlight the execution strategies between detected gestures, as robust execution requires additional control strategies. A gesture becomes actionable only after exceeding a minimal dwell time of 0.1 seconds, which suppresses spurious activations from transient contacts. Commands are executed in velocity-control mode, with each recognized gesture mapped to a predefined velocity profile, and terminate automatically upon contact lift-off. At any moment, only one gesture remains active; a new gesture preempts the current one once the dwell-time condition is satisfied. All other tactile inputs, such as undefined gestures or the absence of contact, are treated as invalid gestures and do not trigger robot actions. 

\section{Gesture Recognition}
\label{sec:gesture_recognition}

\begin{figure*}[t]
\centering
\includegraphics[width=\textwidth]{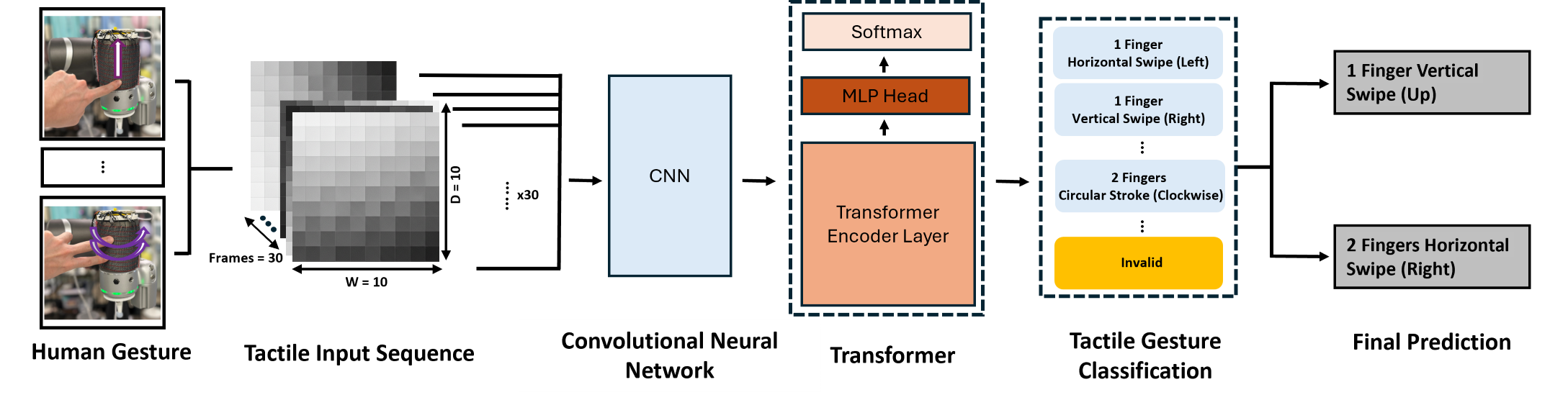} 
\caption{Overview of the proposed gesture recognition pipeline. From left to right: human gestures applied on the tactile skin generate spatiotemporal signals, forming the tactile input sequences. These sequences are then processed by our hybrid CNN–Transformer network to perform tactile gesture classification and produce the final prediction results.}
\label{fig:architecture}
\end{figure*}

Gestures applied on the tactile skin produce spatiotemporal signals, where each time step is represented by a 2D tactile frame on the $10 \times 10$ sensing grid. A sequence of $30$ frames corresponds to a $150~ms$ gesture window, which we empirically found sufficient to capture the shortest meaningful gestures while maintaining smooth and real-time interaction. We formulate the recognition objective as classifying each sequence into one of $15$ gesture classes, comprising $14$ predefined gestures and $1$ invalid class otherwise.

Conventional 3D CNNs can capture spatiotemporal correlations, but their temporal receptive field grows only with kernel size and depth, making them inefficient for modeling long-range dependencies in real time. Recurrent models such as LSTMs preserve temporal order but flatten each 2D frame into a one-dimensional sequence, discarding its spatial structure. 
To balance efficiency and accuracy, we adopt a hybrid architecture that includes a lightweight convolutional stem with a temporal transformer, as illustrated in Fig.~\ref{fig:architecture}. Each tactile frame $x_t \in \mathbb{R}^{10 \times 10}$ is firstly encoded by the convolutional stem into an embedding $z_t \in \mathbb{R}^{64}$. A gesture sequence of $30$ frames is thus represented as $Z \in \mathbb{R}^{30 \times 64}$, which is augmented with positional encodings and processed by $4$ transformer encoder layers to capture temporal dependencies while preserving spatial correlations. Then the encoded features are fed to an MLP head with softmax activation to yield the final classification results over 15 gesture classes.

\begin{figure}
    \centering
    \includegraphics[width=1\columnwidth]{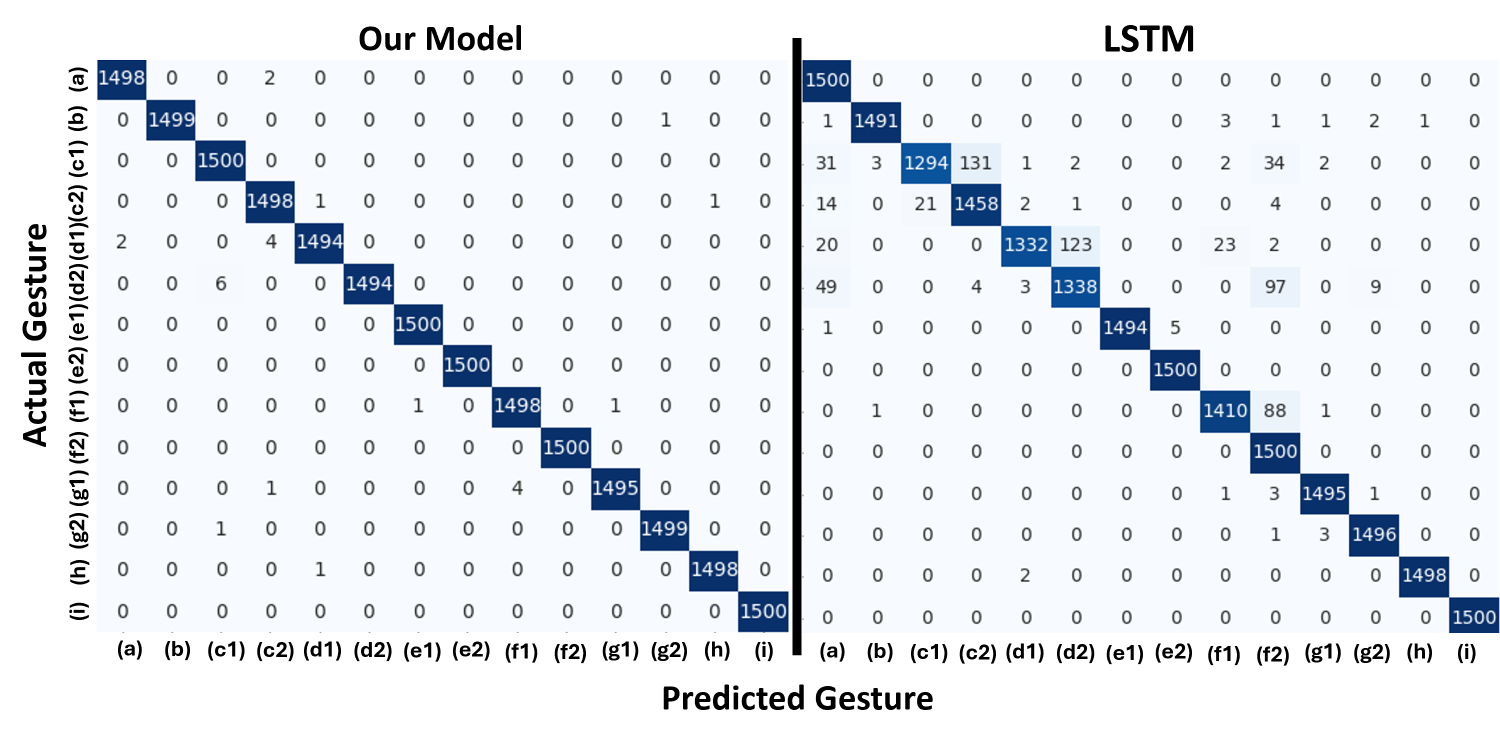}
    \caption{Confusion matrix comparison of our proposed hybrid architecture model (left) and the LSTM baseline (right) on the validation set. Gesture labels correspond to the definitions shown in Fig.~\ref{fig:gesture_design}.}
    \label{fig:confusion_matrix}
\end{figure}

To enhance the model's generalization and prevent the model from overfitting the training data, each original recording of data was used to generate 1,000 samples, each of which contains 30 frames. The data cleaning process involved slicing random-length temporal segments from the core gesture and overlaying them at random temporal positions onto a canvas of background noise. As a result, our full dataset comprises 140,000 augmented gesture sequences, from which $85\%$ and $15\%$ were split as the training and validation sets, respectively. 

As shown in Fig.~\ref{fig:confusion_matrix}, the proposed gesture recognition model achieved an outstanding overall classification accuracy of near-$100\%$, as evidenced by the confusion matrices. 
The matrices, with each x and y coordinate representing an individual gesture, reveal a remarkably clean diagonal indicating that misclassifications are extremely rare with our trained model, and can perfectly differentiate between simple directional and rotational swipes while also classifying complex multi-touch actions with near-perfect accuracy. Compared to a bidirectional LSTM baseline trained on the same dataset, which achieves a total classification accuracy of 96.7\%, our hybrid network demonstrates greater robustness. Although our model’s average inference time is slightly higher at 15 ms per sequence—compared to 12 ms for the LSTM—it still meets the requirements for real-time interaction.

\section{Experiments and results}
\label{sec:experiments}

This section validates the performance of the proposed human-robot interaction method. The detailed experiment process can also be found in the supplemental video.

\subsection{Experimental Setup}
We conducted physical experiments on a 6-DOF robot arm, TM5-900 robot, to evaluate the effectiveness of the proposed gesture-driven control framework. Two representative manipulation tasks were selected: pick-and-place and pouring. The tactile skin was mounted on the robot’s end link, placing the interaction interface directly at the site of manipulation. The skin was affixed using a flexible adhesive layer and wrapped around the robot’s end link, ensuring stable attachment while conforming to its curved geometry.

For comparison, we evaluated our gesture-driven interface against two baselines. The first baseline used a keyboard-based control interface that we developed in-house, where each key was explicitly mapped to the same action defined in the gesture vocabulary. The second baseline employed the manufacturer-provided teach pendant with its graphical user interface (GUI), which also has the same action. We invite a total of $6$ participants ($3$ novices and $3$ experts) to participate in the experiments, enabling the evaluation of effectiveness across different expertise levels.

\subsection{Task Design}
In collaborative robotics, the effectiveness of human–robot interaction is not only measured by task success but also by how intuitively humans can guide the robot to complete actions. To evaluate this, we designed two representative manipulation tasks: pick-and-place and pouring. Both tasks involve direct physical cooperation between the human user and the robot arm, where the quality of interaction critically affects execution efficiency.

\subsubsection{Pick-and-Place}
In this task, the robot was instructed to grasp an object from a fixed location and move it to a designated target position, shown in Fig.~\ref{fig:experimental_setups}(a). In a visionless setting, where the robot lacks autonomous perception of its surroundings, successful execution necessarily relies on human guidance through direct interaction.

\subsubsection{Pouring}
\begin{figure}[t]
    \centering
    
    \begin{subfigure}[b]{0.48\columnwidth}
        \centering
        \includegraphics[width=\textwidth]{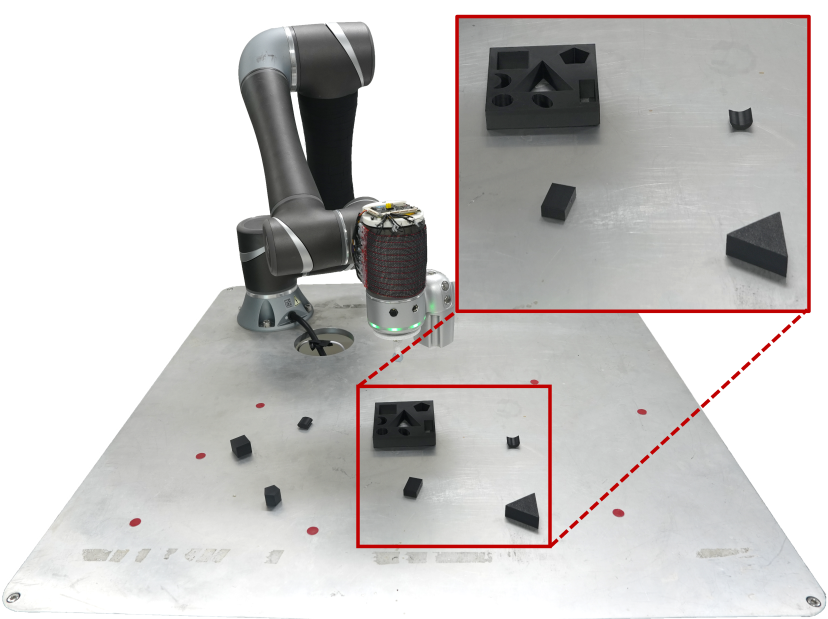}
        \caption{} 
        \label{fig:pick_place}
    \end{subfigure}
     \hfill
    \begin{subfigure}[b]{0.48\columnwidth}
        \centering
        \includegraphics[width=\textwidth]{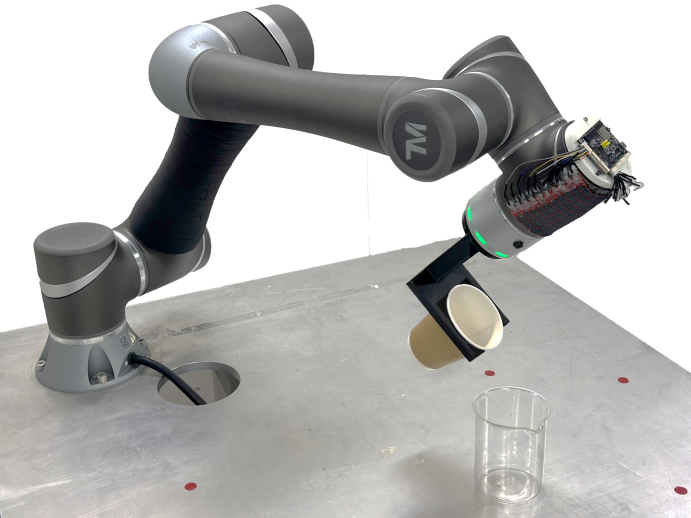}
        \caption{} 
        \label{fig:pouring}
    \end{subfigure}
    
    \caption{Experiment setups. (a) A pick-and-place task in which various 3D-printed shapes need to be placed into their corresponding molds. (b) A pouring task, in which the robot arm pours water from a cup into another container.}
    \label{fig:experimental_setups}
\end{figure}

The pouring task instructed the robot arm to lift a cup, tilt it to pour liquid into a receiving container Fig.~\ref{fig:experimental_setups}(b). In this task, the quality of motion becomes particularly important: trajectories must remain smooth, stable, and consistent to ensure that liquid flow is controlled throughout the pouring process.

\subsection{Evaluation and Comparison}
We focused on primary evaluation metrics as follows.
\begin{enumerate}
    \item \textbf{Completion time}, defined as the total time from the initiation of a command to task completion. Shorter completion times indicate that the interface allows users to express intent more efficiently.
    \item \textbf{Command count}, defined as the total number of discrete control inputs (gestures, key presses, or GUI operations) executed during a trial. Lower counts indicate that users could translate their intent into robot motion more directly, with fewer redundant or corrective actions.
    \item \textbf{Success rate} - this applies to the pouring task, which is defined as the percentage of trials where liquid was poured into the target container without spillage. This metric captures the motion quality of the robot.
\end{enumerate}

Table~\ref{tab:pickplace} and Table~\ref{tab:pouring} present the averaged results of the pick-and-place and pouring tasks across the three control interfaces. Each participant performed $10$ trials per task, and values are averaged over all trials and participants; standard deviations are also reported. We report three metrics: \emph{Time (s)} for task completion, \emph{Commands} for the number of discrete inputs. For pouring, \emph{Success Rate (SR \%)} indicates the percentage of trials completed without spillage.

Across both tasks, the tactile–gesture interface yielded shorter completion times than the two baselines for both novices and experts. The relative reduction was larger for novices (35–57\%) than for experts (23–50\%), indicating a lower entry barrier for novice users; experts also showed consistent but smaller gains. Command counts with gesture control were also lower, showing that participants could convey intent with fewer redundant actions. Additionally, in the pouring task, which requires finer-grained motion control, gesture-driven control is naturally mapped to robot motion, resulting in smoother robot trajectories with fewer interruptions. This allowed participants to complete with a higher success rate.

Overall, these results demonstrate that the capacitance-based woven tactile skin, when employed as an embodied interface, improves the efficiency of robot control and highlights its potential to advance intuitive human–robot interaction by aligning robot actions more closely with natural human intent.

\begin{table}[t]
\centering
\caption{Statistics results on the Pick-and-Place task.}
\label{tab:pickplace}
\begin{tabular}{lccc}
\hline
\textbf{Interface} & \textbf{Time (s)} & \textbf{Cmds} & \textbf{SR (\%)} \\
\hline
Tactile Skin (Nov.)  & $\mathbf{85.4 \pm 3.4}$ & $\mathbf{12+0.3}$ & -- \\
Tactile Skin (Exp.)  & $\mathbf{72.3 \pm 2.1}$ & $\mathbf{7+0.1}$ & -- \\
Keyboard (Nov.)      & $146.7 \pm 9.5$          & $32+1.7$          & -- \\
Keyboard (Exp.)      & $102.1 \pm 5.1$          & $26+1.2$          & -- \\
Teach Pendant (Nov.) & $182.7 \pm 12.2$          & $38+2.4$          & -- \\
Teach Pendant (Exp.) & $132.4 \pm 7.8$          & $29+2.0$          & -- \\
\hline
\end{tabular}
\end{table}
\vspace{5px}
\begin{table}[t]
\centering
\caption{Statistics results on the pouring task.}
\label{tab:pouring}
\begin{tabular}{lccc}
\hline
\textbf{Interface} & \textbf{Time (s)} & \textbf{Cmds} & \textbf{SR (\%)} \\
\hline
Tactile Skin (Nov.)  & $\mathbf{62.4 \pm 2.1}$ & $\mathbf{11+0.4}$ & $\mathbf{98.7}$ \\
Tactile Skin (Exp.)  & $\mathbf{51.5 \pm 1.9}$ & $\mathbf{8+0.2}$ & $\mathbf{99.2}$ \\
Keyboard (Nov.)      & $105.4 \pm 3.8$          & $18+1.6$          & $94.3$ \\
Keyboard (Exp.)      & $74.3 \pm 2.8$          & $12+1.3$          & $96.6$ \\
Teach Pendant (Nov.) & $132.9 \pm 7.6$          & $23+2.5$          & $91.2$ \\
Teach Pendant (Exp.) & $94.1 \pm 5.3$          & $15+2.1$          & $93.4$ \\
\hline
\end{tabular}
\end{table}

\section{Conclusions and Future Work}

This paper presented an embodied gesture–driven human–robot interaction framework based on a capacitance-based woven tactile skin. Experiment results highlight the potential of woven tactile skins as practical embodied interfaces that align robot actions more closely with natural human intent. Beyond HRI, woven tactile skins also open opportunities for data-driven embodied intelligence: by collecting large-scale embodied interaction data, they can provide rich supervision for learning smooth motion policies, optimal path planning by leveraging human prior knowledge embedded in gesture–action demonstrations. This integration allows robots to acquire skills that generalize across tasks and environments. 

The proposed method still has a large room for improvement. First, the current skin design is limited to developable surfaces; as a planar structure, it cannot yet conform to non-developable curved geometries (general 2-manifolds), which constrains its applicability on complex robot bodies. Second, although the current gesture vocabulary covers essential commands, a broader set of gestures will need to be explored to support richer collaborative tasks.

Future work includes exploring larger-area deployment to capture richer interaction patterns and incorporating multimodal signals such as vision or speech to further enhance intuitiveness.

\section*{Acknowledgements}
This study was supported by the Centre for Perceptual and Interactive Intelligence, a CUHK-led InnoCentre under the InnoHK initiative of the Innovation and Technology Commission of the Hong Kong Special Administrative Region Government.

\bibliographystyle{IEEEtran}
\balance
\bibliography{reference}

\end{document}